\documentclass[fleqn,10pt]{wlscirep}

\usepackage{color}
\usepackage{bbm}
\usepackage{mathrsfs}
\usepackage{amsfonts}
\usepackage{amssymb}
\usepackage{graphicx}
\usepackage{cite}
\usepackage[linesnumbered,ruled,lined]{algorithm2e}
\usepackage{array}
\usepackage{subfigure}
\usepackage{multirow}
\usepackage{amsmath}
\usepackage{bm}
\usepackage{upgreek}
\usepackage{float}
\newcommand{\format}[1]{\text{\bf #1}}

\usepackage{fmtcount}
\usepackage{booktabs,threeparttable}

\title{Fourier ptychographic reconstruction using Poisson maximum likelihood and truncated Wirtinger gradient}

\author[1]{Liheng Bian}
\author[1,*]{Jinli Suo}
\author[2]{Jaebum Chung}
\author[2]{Xiaoze Ou}
\author[2]{Changhuei Yang}
\author[1]{Feng Chen}
\author[1]{Qionghai Dai}
\affil[1]{Department of Automation, Tsinghua University, Beijing 100084, China}
\affil[2]{Department of Electrical Engineering, California Institute of Technology, Pasadena, CA 91125, USA}

\affil[*]{jlsuo@tsinghua.edu.cn}

\begin{abstract}
Fourier ptychographic microscopy (FPM) is a novel computational coherent imaging technique for high space-bandwidth product imaging. Mathematically, Fourier ptychographic (FP) reconstruction can be implemented as a phase retrieval optimization process, in which we only obtain low resolution intensity images corresponding to the sub-bands of the sample's high resolution (HR) spatial spectrum, and aim to retrieve the complex HR spectrum. In real setups, the measurements always suffer from various degenerations such as Gaussian noise, Poisson noise, speckle noise and pupil location error, which would largely degrade the reconstruction. To efficiently address these degenerations, we propose a novel FP reconstruction method under a gradient descent optimization framework in this paper. The technique utilizes Poisson maximum likelihood for better signal modeling, and truncated Wirtinger gradient for error removal. Results on both simulated data and real data captured using our laser FPM setup show that the proposed method outperforms other state-of-the-art algorithms. Also, we have released our source code for non-commercial use.
\end{abstract}
\begin{document}

\flushbottom
\maketitle
%
%
\thispagestyle{empty}

\section*{Introduction}\label{sec:Introduction}

\begin{figure}[h]
\centering
\centerline{\includegraphics[width=\textwidth]{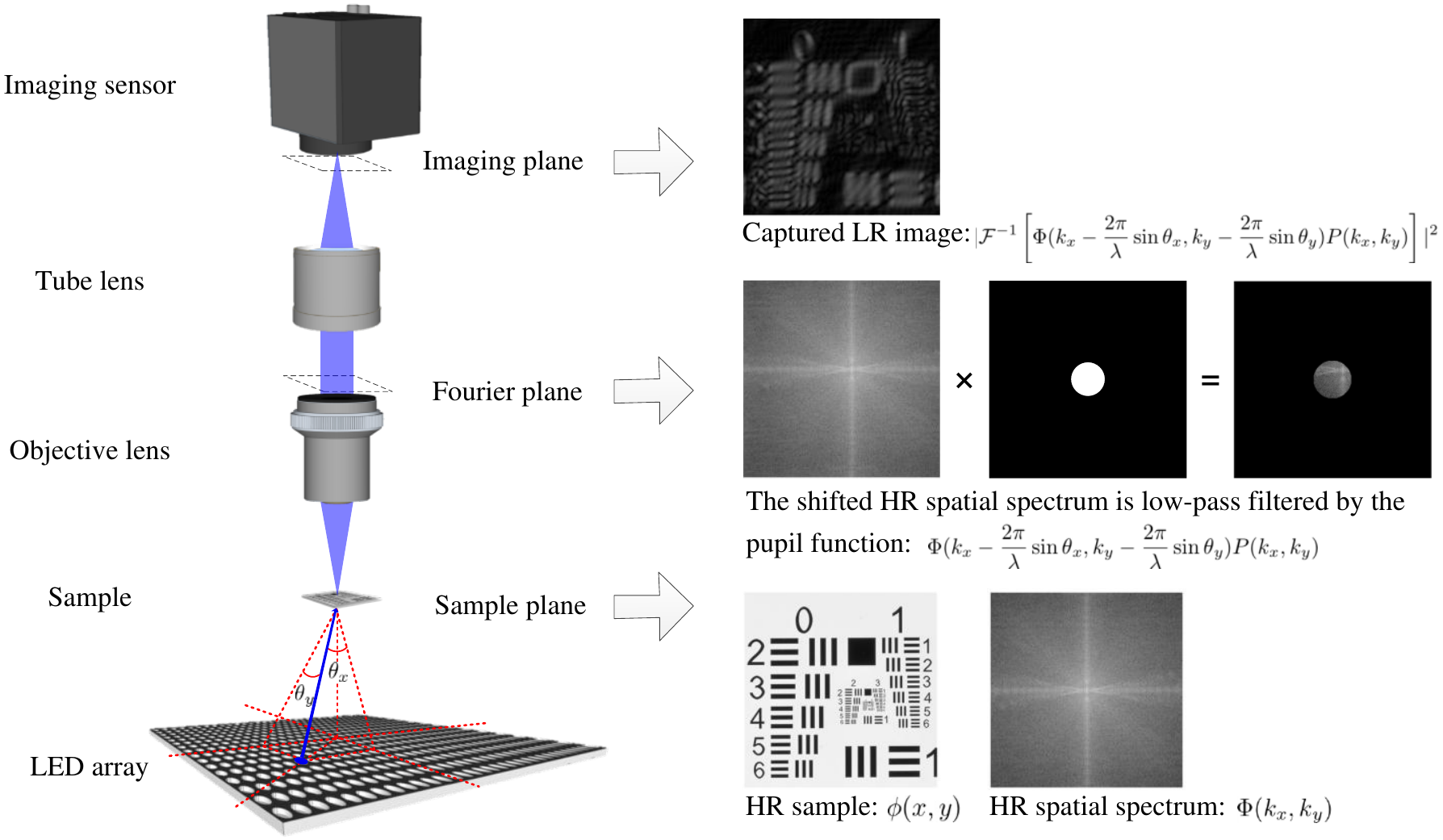}}
\vspace{-1mm}
\caption{The FPM system and its image formation.}
\label{fig:Fig_System}
\end{figure}

Fourier ptychographic microscopy (FPM) is a novel computational coherent imaging technique for high space-bandwidth product (SBP) imaging \cite{FPM_Nature, FPM_Quantitative}. This technique sequentially illuminates the sample with different incident angles, and correspondingly captures a set of low-resolution (LR) images of the sample. Assuming that the incident light is a plane wave and the imaging system is a low-pass filter, the LR images captured under different incident angles correspond to different spatial spectrum bands of the sample, as shown in Fig. \ref{fig:Fig_System}. By stitching these spectrum bands together in Fourier space, a large field-of-view (FOV) and high resolution (HR) image of the sample can be obtained. As a reference, the synthetic NA of the FPM setup in Ref. \cite{FPM_Nature} is $\sim$0.5, and the FOV can reach $\sim$120 mm$^2$, which greatly improves the throughput of the existing microscope. FPM has been widely applied in 3D imaging \cite{3D_1, 3D_2}, fluorescence imaging \cite{Fluo_1, Fluo_2}, mobile microscope \cite{Cellphone_1, Cellphone_2}, and high-speed in vitro imaging \cite{Vitro}.

Mathematically, Fourier ptychographic (FP) reconstruction can be implemented as a typical phase retrieval optimization process, which needs to recover a complex function given the intensity measurements of its linear transforms. Specifically, we only obtain the LR intensity images corresponding to the sub-bands of the sample's HR spatial spectrum, and the reconstruction is to retrieve the complex HR spatial spectrum. Conventional FPM \cite{FPM_Nature, FPM_Quantitative} utilizes the alternating projection (AP) algorithm \cite{Phase_Comparison, Phase_Fienup_2}, which adds constraints alternatively in spatial space (captured intensity images) and Fourier space (pupil function), to stitch the LR sub-spectra together. AP is easy to implement and fast to converge, but is sensitive to measurement noise and system errors. To tackle measurement noise, Bian et al. \cite{FPMAlgorithm_Bian} proposed a novel method termed Wirtinger flow optimization for Fourier Ptychography (WFP), which uses gradient descent scheme and Wirtinger calculus \cite{Phase_Wirtinger} to minimize the intensity errors between estimated LR images and measurements. WFP is robust to Gaussian noise, and can produce better reconstruction results in low-exposure imaging scenarios and thus largely decrease image acquisition time. However, it needs careful initialization since the optimization is non-convex. Based on the semidefinite programming (SDP) convex optimization for phase retrieval \cite{Phase_Lift_2, Phase_Cut}, Horstmeyer et al. model FP reconstruction as a convex optimization problem \cite{FPMAlgorithm_Convex}. The method guarantees a global optimum, but converges slow which makes it impractical in real applications. Recently, Yeh et al. \cite{FPMAlgorithm_Laura} tested different objective functions (intensity based, amplitude based and Poisson maximum likelihood) under the gradient-descent optimization scheme for FP reconstruction. The results show that the amplitude-based and Poisson maximum-likelihood objective functions produce better results than the intensity-based objective function. To address the LED misalignment, the authors also add a simulated annealing algorithm into each iteration to search for optimal pupil locations.

Although the above methods offer various options for FP reconstruction, they have their own limitations. AP and WFP are limited to adress Gaussian noise, and cannot handle speckle noise well (as shown in following experiments) which is common when the light source is highly spatially and temporally coherent  (such as laser) \cite{LaserFPM}. 
The Poisson Wirtinger Fourier ptychographic reconstruction (PWFP) technique mentioned in Ref.\cite{FPMAlgorithm_Laura} performs better than other methods, but it still obtains aberrant reconstruction results with measurements corrupted with Gaussian noise, and needs much more running time for the incorporated simulated annealing algorithm to deal with LED misalignment.

In this paper, we propose a novel FP reconstruction method termed as truncated Poisson Wirtinger Fourier ptychographic reconstruction (TPWFP), to efficiently handle the above mentioned measurement noise and pupil location error that are common in FPM setups. The technique incorporates Poisson maximum likelihood objective function and truncated Wirtinger gradient \cite{TWF} together into a gradient-descent optimization framework. The advantages of TPWFP lie in three aspects:
\begin{itemize}
  \item The utilized Poisson maximum-likelihood objective function is more appropriate to describe the Poisson characteristic of the photon detection by an optical sensor in real imaging systems, and thus can obtain better results in real applications.
  \item Truncated gradient is used to prevent outliers from degrading the reconstruction, which provides better descent directions and enhanced robustness to various sources of error such as Gaussian noise and pupil location error.
    \item There is no matrix lifting and global searching for optimization, resulting in faster convergence and less computational requirement.
\end{itemize}
To demonstrate the effectiveness of TPWFP, we test it against the aforementioned algorithms on both simulated data and real data captured using a laser FPM setup \cite{LaserFPM}. Both simulations and real experiments show that TPWFP outperforms other state-of-the-art algorithms in imaging scenarios involving Poisson noise, Gaussian noise, speckle noise and pupil location error.

\section*{Methods}

As stated before, TPWFP incorporates Poisson maximum likelihood objective function and truncated Wirtinger gradient together into a gradient descent optimization framework for FP reconstruction. Next, we begin to introduce this technique in detail.

\subsection*{Image formation of FPM}

FPM is a coherent imaging system. Under the assumption that the light incident on a sample is a plane wave, we can describe the light field transmitted from the sample as $\phi(x, y) e^{(jx\frac{2\pi}{\lambda}\sin\theta_x, jy\frac{2\pi}{\lambda}\sin\theta_y)}$, where $\phi$ is the sample's complex spatial map, $(x, y)$ are the 2D spatial coordinates, $j$ is the imaginary unit, $\lambda$ is the wavelength of illumination, and $\theta_x$ and $\theta_y$ are the incident angles as shown in Fig. \ref{fig:Fig_System}. 
Then the light field is Fourier transformed to the pupil plane when it travels through the objective lens, and subsequently low-pass filtered by the aperture. This process can be represented as $P(k_x, k_y)\mathcal{F}(\phi(x, y) e^{(jx\frac{2\pi}{\lambda}\sin\theta_x, jy\frac{2\pi}{\lambda}\sin\theta_y)})$, where $P(k_x, k_y)$ is the pupil function for low-pass filtering, $(k_x, k_y)$ are the 2D spatial frequency coordinates in the pupil plane, and $\mathcal{F}$ is the Fourier transform operator. Afterwards, the light field is Fourier transformed again when it passes through the tube lens to the imaging sensor. Since real imaging sensors can only capture light's intensity, the image formation of FPM follows:
\begin{eqnarray}\label{eqs:Formation_0}
\format I &=& |\mathcal{F}^{-1}\left[P(k_x, k_y)\mathcal{F}(\phi(x, y) e^{(jx\frac{2\pi}{\lambda}\sin\theta_x, jy\frac{2\pi}{\lambda}\sin\theta_y)})\right]|^2\\ \nonumber
&=& |\mathcal{F}^{-1}\left[P(k_x, k_y)\Phi(k_x - \frac{2\pi}{\lambda}\sin\theta_x, k_y - \frac{2\pi}{\lambda}\sin\theta_y) \right]|^2,
\end{eqnarray}
where $\format I$ is the captured image, $\mathcal{F}^{-1}$ is the inverse Fourier transform operator, and $\Phi$ is the spatial spectrum of the sample. Visual explanation of the image formation process is diagrammed in Fig. \ref{fig:Fig_System}



Because $\mathcal{F}^{-1}$ is linear and $P(k_x, k_y)\Phi(k_x - \frac{2\pi}{\lambda}\sin\theta_x, k_y - \frac{2\pi}{\lambda}\sin\theta_y)$ is a linear operation of low-pass filtering to the HR spatial spectrum with the pupil function, we rewrite the above image formation of FPM as
\begin{eqnarray}\label{eqs:Formation}
\format b = |\format A\format z|^2,
\end{eqnarray}
where $\format b\in \mathbb{R}^m$ is the ideal captured images (all the captured $\format I$ under different angular illumination in vector form), $\format A\in \mathbb{C}^{m\times n}$ is the corresponding linear transform matrix incorporating Fourier transforms and the low-pass filter, and $\format z\in \mathbb{C}^n$ is the sample's HR spectrum ($\Phi$ in vector form). This is a standard phase retrieval problem \cite{Phase_Review}, where $\format b$ and $\format A$ are known, and $\format z$ is what we aim to recover.

\subsection*{Poisson maximum-ikelihood objective function}

Here we assume that the detected photons at each detector unit follow Poisson distribution in real setups, which is consistent with the independent nature of random individual photon arrivals at the imaging sensor \cite{Poisson}. Note that although the Poisson distribution approaches a Gaussian distribution for large photon counts according to the central limit theorem, most of the captured images in FPM are dark-field images captured under oblique illuminations and thus are more consistent with Poisson distribution.
In a nutshell, the signal's probability model can be represented as
\begin{eqnarray}
c_i\sim Poisson(b_i), ~~~~~~i = 1,\cdots,m,
\end{eqnarray}
where $b_i = |\format a_i \format z|^2$ is the $i$ th latent signal in $\format b$, and $\format a_i$ is the $i$ th row of $\format A$.
Thus, for $c_i$, its probability mass function is
\begin{eqnarray}
g(c_i) &=& \frac{e^{-b_i}b_i^{c_i}}{c_i!},
\end{eqnarray}
where $e$ is the Euler's number, and $c_i!$ is the factorial of $c_i$.

Based on the maximum-likelihood estimation theory, assuming that the measurement are independent from each other, the reconstruction turns into maximizing the global probability of all the measurements $c_{1,\cdots, m}$, namely
\begin{eqnarray}
\max ~~ \prod_{i=1}^{m}g(c_i).
\end{eqnarray}
Taking a logarithm of the objective function yields
\begin{eqnarray}
\min ~~~~ L &=& -log\prod_{i=1}^{m}g(c_i)\\ \nonumber
&=& -log\prod_{i=1}^{m}\frac{e^{-b_i}b_i^{c_i}}{c_i!}\\ \nonumber
&=& -\sum_{i=1}^{m}log(\frac{e^{-b_i}b_i^{c_i}}{c_i!})\\ \nonumber
&=& \sum_{i=1}^{m}b_i - \sum_{i=1}^{m}c_ilog(b_i) + \sum_{i=1}^{m}log(c_i!).
\end{eqnarray}
Since $c_{1,\cdots, m}$ are the experimental measurements and are thus constant in the optimization process, we can omit the last item in $L$ (namely $\sum_{i=1}^{m}log(c_i!)$) for optimization. Then by replacing $b_i$ with $|\format a_i \format z|^2$, we obtain the objective function of TPWFP as
\begin{eqnarray}
\min ~~~~ L(\format z) &=& \sum_{i=1}^{m}\left[|\format a_i \format z|^2 - c_ilog(|\format a_i \format z|^2)\right].
\end{eqnarray}

\subsection*{Truncated Wirtinger gradient}

As stated before, we use the gradient-descent optimization scheme. Based on the Wirtinger calculus \cite{Phase_Wirtinger}, we obtain the gradient of $L(\format z)$ as
\begin{eqnarray}\label{eqs:gradient}
\nabla L(\format z) &=& \frac{d L(\format z)}{d \format z^*} \\ \nonumber
&=& \frac{d \sum_{i=1}^{m}\left[|\format a_i \format z|^2 - c_ilog(|\format a_i \format z|^2)\right]}{d \format z^*} \\ \nonumber
&=& \sum_{i=1}^{m}\frac{d \left[|\format a_i \format z|^2 - c_ilog(|\format a_i \format z|^2)\right]}{d \format z^*}\\ \nonumber
&=& \sum_{i=1}^{m}2\left(\format a_i^H\format a_i\format z - \frac{c_i}{|\format a_i\format z|^2}\format a_i^H\format a_i\format z\right)\\ \nonumber
&=& \sum_{i=1}^{m}2\left(\format a_i\format z - \frac{c_i\format a_i\format z}{|\format a_i\format z|^2}\right)\format a_i^H, \\ \nonumber
\end{eqnarray}
where $\format a_i^H$ is the transposed-conjugate matrix of $\format a_i$.

To prevent optimization degeneration from measurement noise, we add a thresholding operation to $\nabla L(\format z)$ before using it to update $\format z$ in each iteration. Similar to Ref. \cite{TWF}, the thresholding constraint is defined as
\begin{eqnarray}\label{eqs:thresholding}
|c_i - |\format a_i\format z|^2| \leq a^h\frac{||\format c - |\format A\format z|^2||_1}{m}\frac{|\format a_i\format z|^2}{||\format z||}.
\end{eqnarray}
Here $a^h$ is a predetermined parameter specified by users, $|c_i - |\format a_i\format z|^2|$ is the difference between the $i$ th measurement and its reconstruction, $\frac{||\format c - |\format A\format z|^2||_1}{m}$ is the mean of all the differences, and $\frac{|\format a_i\format z|^2}{||\format z||}$ is the relative value of the latent signal. Intuitively, the thresholding indicates that if one measurement is far from the reconstruction, it is labeled as an outlier and omitted in subsequent optimization. Note that the thresholding is signal dependent, which is beneficial for accurate detection of outliers.

Thus, for each index $i$, its corresponding measurement can be used in Eqs. (\ref{eqs:gradient}) only when it meets the thresholding criterion in Eq. (\ref{eqs:thresholding}). 
In the following, we use $\xi$ to denote the index set that meets the thresholding constraint, and represent the corresponding truncated Wirtinger gradient as
\begin{eqnarray}\label{eqs:Tgradient}
\nabla L_{\xi}(\format z) &=& \sum_{i\in \xi}^{}2\left(\format a_i\format z - \frac{c_i\format a_i\format z}{|\format a_i\format z|^2}\right)\format a_i^H. \\ \nonumber
\end{eqnarray}
Note that the index set $\xi$ is iteration-variant, meaning that in each iteration we update $\xi$ according to the measurements $\format c$ and the updated $\format z$. This adaptively provides us with better descent direction.

In the gradient descent scheme, we update $\format z$ in the $k$th iteration as
\begin{eqnarray}\label{eqs:update}
\format z^{(k+1)} = \format z^{(k)} - \mu\nabla L_{\xi}(\format z),
\end{eqnarray}
where $\mu$ is the gradient descent step that is predetermined manually. Here we utilize a setting similar to Ref. \cite{FPMAlgorithm_Bian} as 
\begin{eqnarray}\label{eqs:mu}
\mu^{(k)} = \frac{\min\left(1-e^{-k/k_0},~\mu_{max}\right)}{m},
\end{eqnarray}
where we set $k_0 =  330$ and $\mu_{max} = 0.1$. This type of gradient-descent step is widely used, since it allows the gradient descent step to grow gradually, and offers an adaptive way for better convergence \cite{Phase_Wirtinger, TWF}.

\begin{algorithm}[!th]
\SetKwInOut{Majorization}{Majorization}\SetKwInOut{Minimization}{Minimization}
\SetKwData{set}{set}
\SetKwInOut{Initialization}{Initialization}\SetKwInOut{Input}{Input}\SetKwInOut{Output}{Output}
\vspace{2mm}
\Input{Linear tranform matrix $\bf A$, measurement vector $\format c$, and initialization $\format z^{(0)}$.}
\Output{Retrieved complex signal $\bf z$ (sample's HR spatial spectrum).}
\vspace{2mm}
 $k = 0$\;
      \While{not converged}{
          Update $\xi$ according to Eq. (\ref{eqs:thresholding})\;
          Update $\mu^{(k+1)}$ according to Eq. (\ref{eqs:mu})\;
          Update $\format z^{(k+1)} $ according to Eq. (\ref{eqs:update})\;
          $k:=k+1$.
          }
\caption{\small{TPWFP algorithm for FP reconstruction}}
\label{alg:TPWFP}
\end{algorithm}

Based on the above derivations, we summarize the proposed TPWFP algorithm in Alg. \ref{alg:TPWFP}. For the initialization $\format z^{(0)}$, similar to Ref. \cite{FPMAlgorithm_Bian}, we set $\format z^{(0)}$ as the spatial spectrum of the up-sampled version of the LR image captured under the normal incident light. According to Ref. \cite{TWF}, the computation complexity of such an optimization algorithm is $\mathcal{O}(mn \log \frac{1}{\epsilon})$, where $m$ is the number of measurements, $n$ is the number of signal entries, and $\epsilon$ is the relative reconstruction error defined in Eq. (\ref{eqs:re}). This is much lower than WFP's computation complexity which is $\mathcal{O}(mn^2 \log \frac{1}{\epsilon})$. 
Note that the source code of TPWFP is available at \href{http://www.sites.google.com/site/lihengbian}{\textit{http://www.sites.google.com/site/lihengbian}} for non-commercial use.

\section*{Results}

In this section, we test the proposed TPWFP and other state-of-the-art algorithms including AP, WFP and PWFP on both simulated and real captured data, to show pros and cons of each method.

\subsection*{Quantitative metric}

To quantitatively evaluate the reconstruction quality, we utilize the relative error (RE) \cite{TWF} metric defined as
\begin{equation}\label{eqs:re}
RE(\textbf{z}, \hat{\textbf{z}}) = \frac{\min_{\phi \in [0, 2\pi)}||e^{-j\phi}\textbf{z} - \hat{\textbf{z}}||^2}{||\hat{\textbf{z}}||^2}.
\end{equation}
This metric describes the difference between two complex functions $\format z$ and $\hat{\textbf{z}}$. We use it here to compare reconstructed HR spatial spectrum with groundtruth in the simulation experiments.

\subsection*{Parameters}

In the simulation experiments, we simulate the FPM setup with its hardware parameters as follows: the numerical aperture (NA) of the objective lens is 0.08, and the corresponding pupil function is an ideal binary function (all ones inside the NA circle and all zeros outside); the height from the LED plane to the sample plane is 84.8mm; the distance between adjacent LEDs is 4mm, and $15\times 15$ LEDs are used to provide a synthetic NA of $\sim$0.5; the wavelength of incident light is 625nm; and the pixel size of captured images is 0.2um. Besides, we use the 'Lena' and the 'Aerial' image ($512\times512$ pixels) from the USC-SIPI image database \cite{Data} as the latent HR amplitude and phase map, respectively. The captured LR images' pixel numbers are set to be one tenth of the HR image along both dimensions, and are synthesized based on the image formation in Eq. \ref{eqs:Formation}. We repeat 20 times for each of the following simulation experiments and average their evaluations to produce the final results.

\begin{figure}[h]
\centering
\centerline{\includegraphics[width=\textwidth]{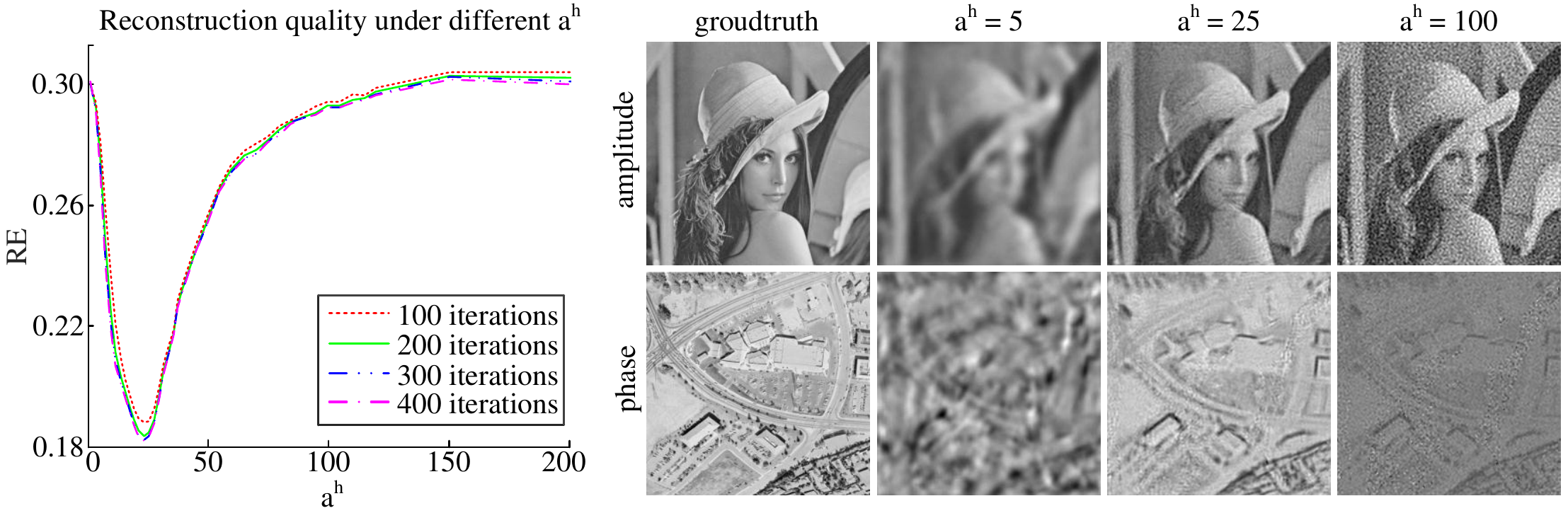}}
\vspace{-1mm}
\caption{Reconstruction quality (relative error between reconstructed HR spatial spectrum and groundtruth) under different settings of $a^h$ in TPWFP. The experiment is conducted on the simulated dataset corrupted with white Gaussian noise (std = $2\times10^{-3}$).}
\label{fig:Fig_Thresholding}
\end{figure}

As for the algorithms' parameters, an important parameter of TPWFP is the thresholding $a^h$ in Eq. \ref{eqs:thresholding}. To choose  appropriate $a^h$, we test different $a^h$ on the simulated data corrupted with white Gaussian noise, and study its influence on the final reconstruction quality. The results are shown in Fig. \ref{fig:Fig_Thresholding}, from which we can see that both too small or too big of $a^h$ result in worse reconstruction. This is determined by the nature of the utilized truncated gradient. When $a^h$ is too small, more informative measurements are incorrectly labeled as outliers and thus contribute nothing to final reconstruction, resulting in blurred reconstructed images as shown in Fig. \ref{fig:Fig_Thresholding}. When $a^h$ is too big, measurement noise and system errors are not effectively removed from reconstructed images. To sum up, we choose an appropriate assignment for $a^h$ as $a^h = 25$ in the following experiments for TPWFP.

Another parameter for all the algorithms is the iteration number. For AP, we set its iteration number to be 100 which is enough for its convergence \cite{FPM_Nature}. For WFP, its iteration number is set to be 1000 \cite{FPMAlgorithm_Bian}. From Fig. \ref{fig:Fig_Thresholding} we can see that 200 iterations are enough for TPWFP to converge. Since PWFP is a particular form of TPWFP when $a^h = \infty$ (no thresholding to the gradient), we set the same iteration number for PWFP.

\subsection*{Simulation experiments}\label{sec:SimulationExperiments}

\begin{figure}[!t]
\centering
\centerline{\includegraphics[width=0.85\textwidth]{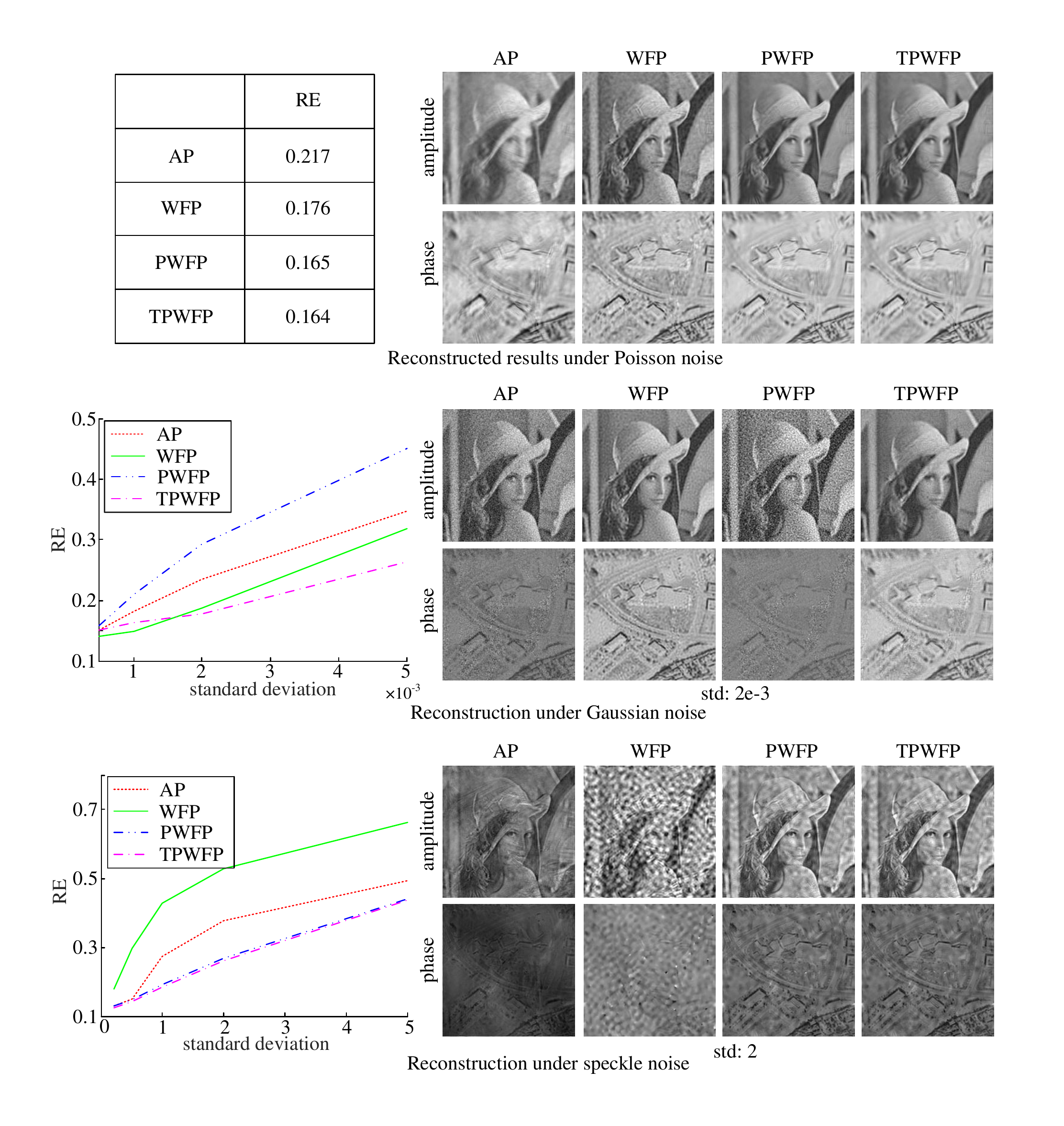}}
\caption{Comparison of the reconstruction results by the three state-of-the-arts and the proposed TPWFP under Poisson noise, Gaussian noise and speckle noise.}
\label{fig:Fig_Noise}
\end{figure}

First, we test the four algorithms (AP, WFP, PWFP and TPWFP) on simulated captured images corrupted with Poisson noise, Gaussian noise and speckle noise, respectively, which are the most common noise in real imaging setups. The first two kinds are caused by photoelectric effect and dark current \cite{Poisson}, while speckle noise are caused by the light's spatial-temporal coherence of the illumination source (such as laser).

The results are shown in Fig. \ref{fig:Fig_Noise}. Note that the standard deviation (std) is the ratio between actual std and the maximum of the ideal measurements $\format b$. Besides, we use the model $\format c = \format b(1+\format n)$ for speckle noise simulation, where $\format n$ is uniformly distributed random zero mean noise. 
From the results, we can see that under small Gaussian noise, WFP outperforms the other three methods. This is because WFP assumes global uniform noise for both low and high spatial frequencies, which is consistent with the corrupted noise model. Instead, PWFP and TPWFP assume that noise would be smaller for lower intensities (especially for LR images correspond to high spatial frequencies). Thus, they cannot remove enough noise for these spatial frequencies and produce worse results. However, when noise grows to around std$\geqslant 2\times 10^{-3}$, TPWFP get better results than WFP. This is because when noise is large, WFP cannot extract useful information from the noisy data, while TPWFP directly omits these measurements to avoid their negative influence on final reconstruction. For Poisson noise and speckle noise, while both PWFP and TPWFP obtains better results than the other methods, TPWFP is little advantageous than PWFP. This is because for these kinds of signal dependent noise, it is hard for the truncated gradient to correctly distinguish noise from latent signals.

\begin{figure}[!ht]
\centering
\centerline{\includegraphics[width=0.9\textwidth]{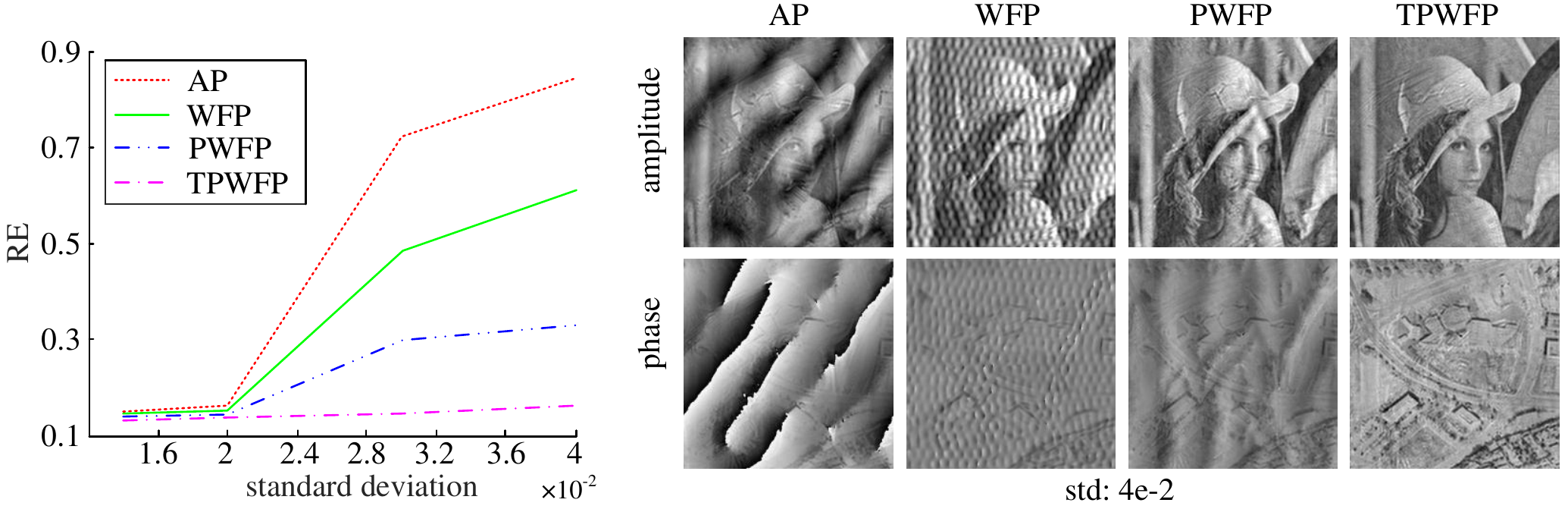}}
\caption{Reconstruction results by the three state-of-the-arts and the proposed TPWFP under pupil location error.}
\label{fig:Fig_Pupil}
\end{figure}

Then we apply the four algorithms on the simulated data corrupted with pupil location error, which is common in real setups due to LED misalignment or unexpected system errors. We simulate pupil location error by adding Gaussian noise to the incident wave vectors of each LED. The reconstruction results are shown in Fig. \ref{fig:Fig_Pupil}. From the results we can see that TPWFP outperforms state-of-the-arts a lot. This benefits from the nature of the utilized truncated gradient. In the thresholding operation (Eq. (\ref{eqs:thresholding})), if one measurement (spatial space) is far from reconstruction due to pupil location error, we omit this measurement which represents misaligned information. Thus, we prevent the pupil location error from degenerating final reconstruction.

\begin{table*}[!h]
\centering
\caption{Comparison of running time between state-of-the-arts and the proposed TPWFP.}\label{tab:RunningTime}
{\small
\begin{tabular}{p{0.2\linewidth}<{\centering}p{0.15\linewidth}<{\centering}p{0.15\linewidth}<{\centering}p{0.15\linewidth}<{\centering}p{0.15\linewidth}<{\centering}}
\cline{1-5}
& AP & WFP & PWFP & TPWFP \\
\cline{1-5}
~Iteration & 100    &  1000    & 200 & 200\\
\cline{1-5}
~Running time (s) & 37 & 301 & 65 &  117\\
\cline{1-5}
\end{tabular}
}
\end{table*}

We display the running time of each method under current algorithm settings in Table \ref{tab:RunningTime}. All the four methods are implemented using Matlab on an Intel Xeon 2.4 GHz CPU computer, with 8G RAM and 64 bit Windows 7 system.
From the table we can see that both PWFP and TPWFP save more running time than WFP due to faster convergence \cite{TWF}, but they are still more time consuming than AP. Note that TPWFP consumes more time than PWFP, which is caused by the additional thresholding and truncation operation to the gradient.

\subsection*{Real experiment}

\begin{figure}[!h]
\centering
\centerline{\includegraphics[width=\textwidth]{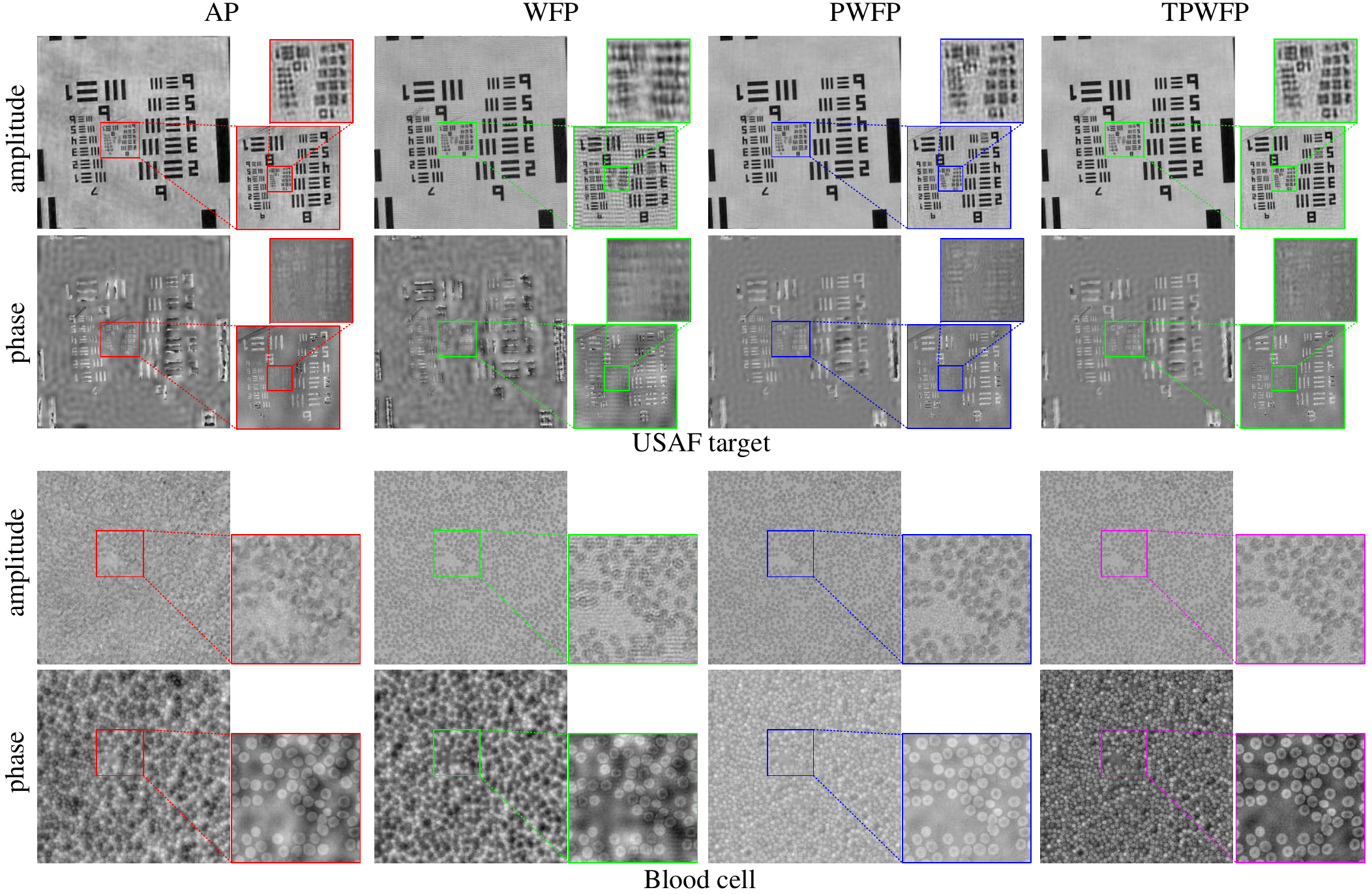}}
\vspace{-1mm}
\caption{Reconstruction results by the three state-of-the-arts and the proposed TPWFP under real captured dataset (USAF target and red blood cell) using our laser FPM setup.}
\label{fig:Fig_Real}
\end{figure}

To further validate the robustness of TPWFP to the above measurement noise and system errors, we run the four algorithms on two real captured datasets including USAF target and red blood cell sample using a laser FPM setup \cite{LaserFPM}. The red blood cell sample is prepared on a microscope slide stained with Hema 3 stain set (Wright-Giemsa). The setup consists of a $4f$ microscope system with a $4\times$ 0.1 NA objective lens (Olympus), a 200 mm focal-length tube lens (Thorlabs), and a 16-bit sCMOS camera (PCO.edge 5.5). The system is fitted with a circular array of 95 mirror elements providing illumination NA of 0.325, resulting in the total synthetic NA of 0.425. A 1W laser of 457 nm wavelength is used for the illumination source, which is pinhole-filtered, collimated and guided to a pair of Galvo mirrors (Thorlabs GVS212) to be directed to individual mirror elements. The reconstruction results are shown in Fig. \ref{fig:Fig_Real}. 
From the results we can see that AP produces intensity fluctuations in the background of reconstructed images (see the white background of the USAF target for clear comparison) and low image contrast (see the reconstructed amplitude of the red blood cell sample). WFP also obtains corrugated artifacts due to the speckle noise produced by the laser illumination. Both PWFP and TPWFP obtain better results than AP and WFP, while TPWFP produces results with more image details (see the reconstructed amplitude of the USAF target, especially group 10) and image contrast (see the reconstructed phase of the red blood cell sample) than PWFP. To conclude, TPWFP outperforms the other methods with less artifacts, higher image contrast and more image details.

\section*{Discussion}

In this paper, we propose a novel reconstruction method for FPM termed as TPWFP, which utilizes Poisson maximum likelihood objective function and truncated Wirtinger gradient for optimization under a gradient descent framework. Results on both simulated data and real data captured using our laser FPM setup show that the proposed method outperforms other state-of-the-art algorithms in cases of Poisson noise, Gaussian noise, speckle noise and pupil location error.

TPWFP can be widely extended. First, the pupil function updating procedure of the EPRY-FPM algorithm \cite{PupilFunction} can be incorporated into TPWFP to obtain corrected pupil function and better reconstruction. Besides, since the linear transform matrix $\format A$ can be composed of any kinds of linear operations (Fourier transform and low-pass filtering in FPM), TPWFP can be applied in various linear optical imaging systems for phase retrieval, such as conventional ptychography \cite{Ptychography}, multiplexed FP \cite{FPM_Multiplexing_1, FPM_Multiplexing_2} and fluorescence FP \cite{Fluo_1}. Also, since TPWFP is much more robust to pupil location error than other methods, it may find wide applications in other imaging schemes where precise calibrations are unavailable.

In spite of advantageous performance and wide applications, the limitations of TPWFP lie in two aspects. First, it is still time consuming compared to conventional AP, though it is much faster than WFP. Second, it is non-convex. Although choosing the up-sampled image as the initialization can result in satisfying reconstruction as we demonstrate in the above experiments, there is no theoretical guarantee for its global optimal convergence.


\section*{Acknowledgements}

This work was supported by the National Natural Science Foundation of China, Nos. 61120106003 and 61327902.

\section*{Author contributions statement}

L.B. and J.S. proposed the idea and conducted the experiments. J.C. and X.O. built the setup. All the authors contributed to writing and revising the manuscript, and convolved in discussions during the project.

\section*{Additional information}

The authors declare no competing financial interests.

\end{document}